\documentclass[twoside]{article} 
\usepackage[accepted]{aistats2017}

\usepackage{times}
\usepackage{comment}
\usepackage{algorithm}
\usepackage[normalem]{ulem}
\usepackage{graphicx}
\usepackage{subfigure}
\usepackage{algpseudocode}
\usepackage{amssymb}
\usepackage{amsthm}
\usepackage[fleqn]{amsmath}

\usepackage{color}
\newcounter{bar}

\newcommand{\G}{{\mathcal{G}}}

\newtheorem{defn}{Definition}[section]
\newtheorem{thm}{Theorem}[section]

\newcommand{\bug}
    {\mbox{\rule{2mm}{2mm}}}

\newcommand{\bi}{\begin{itemize}}
\newcommand{\ei}{\end{itemize}}
\newcommand{\BE}{\begin{enumerate}}
\newcommand{\EE}{\end{enumerate}}

\newcommand{\initab}{                           
\begin{tabbing}
XXX \= XXXX \= \kill
}
\newcommand{\begpub}{
\begin{quotation}
\noindent
}

\newcommand{\finpub}{
\end{quotation}
}

\newcommand{\vv}  {VV-Symmetry}
\newcommand{\vvorbital}  {VV-Orbital MCMC}
\newcommand{\necorbital}  {NEC-Orbital MCMC}

\begin{document}

%

%

\twocolumn[

\aistatstitle{
Non-Count Symmetries in Boolean \& Multi-Valued Prob. Graphical Models
}

\aistatsauthor{ Ankit Anand$^1$  \And Ritesh Noothigattu$^1$ \And Parag Singla \and Mausam }


\aistatsaddress{ Department of CSE \\ I.I.T Delhi \And Machine Learning Department\footnotemark  
\\ Carnegie Mellon University  \And Department of CSE \\ I.I.T Delhi} ]

\footnotetext{First two authors contributed equally to the paper. Most of the work was done while the second author was at IIT Delhi.}

\begin{abstract}
Lifted inference algorithms commonly exploit symmetries in a probabilistic graphical model (PGM) for efficient inference. 
However, existing algorithms for Boolean-valued domains can identify only those pairs of states as symmetric, in which the number of ones and zeros match exactly (\emph{count symmetries}). Moreover,  algorithms for lifted inference in multi-valued domains also compute a multi-valued extension of count symmetries only. These algorithms miss many symmetries in a domain.

In this paper, we present first algorithms to compute {\em non-count symmetries} in both Boolean-valued and multi-valued domains. Our methods can also find symmetries between multi-valued variables that have different domain cardinalities. 
The key insight in the algorithms is that they change the unit of symmetry computation from a variable to a variable-value (VV) pair.  Our experiments find that exploiting these symmetries in MCMC  can obtain substantial computational gains over existing algorithms.

\end{abstract} 
\vspace{-0.1in}
\section{Introduction}
\vspace{-0.1in}
A popular approach for efficient inference in probabilistic graphical models (PGMs) is {\em lifted inference} (see \cite{kimmig&al15}), which identifies repeated sub-structures (symmetries), and exploits them for computational gains. Lifted inference algorithms typically cluster symmetric states (variables) together and use these clusters to reduce computation, for example, by avoiding repeated computation for all members of a cluster via a single representative. Lifted versions of several inference algorithms have been developed such as variable elimination \cite{poole03,braz&al05}, weighted model counting \cite{gogate&domingos11}, knowledge compilation \cite{broeck&al11}, belief propagation \cite{singla&domingos08,kersting&al09,singla&al14}, variational inference \cite{bui&al13}, linear programming \cite{noessner&al13,mladenov&al14} and Markov Chain Monte Carlo (MCMC) \cite{venugopal&gogate12,gogate&al12,niepert12,broeck&niepert15,anand&al16}. 


Unfortunately, to the best of our knowledge, all algorithms compute a limited notion of symmetries, which we call {\em count symmetries}. A count symmetry in a Boolean-valued domain is a symmetry between two states where the total number of zeros and ones exactly match. An illustrative algorithm for Boolean-valued PGMs (which we build upon) is Orbital MCMC \cite{niepert12}. It first uses graph isomorphism to compute symmetries and later uses these symmetries in an MCMC algorithm. Symmetries are represented via permutation groups in which variables interchange values to create other symmetric states. Notice, that if a state has $k$ ones then any permutation of that state will also have $k$ ones; this algorithm can only compute count symmetries.

Similarly, lifted inference algorithms for multi-valued PGMs (e.g., \cite{poole03,bui&al13}), only compute a weak extension of count symmetries for multi-valued domains -- they allow symmetries only between those sets of variables that have the same domain. And, the count, i.e. the number of occurences, of any value (from the domain) within this set of variables remains the same between two symmetric states.




In response, we develop extensions to existing frameworks to enable computation of {\em non-count symmetries} in which the count of a value between symmetric states can change. We can also compute a special form of non-count symmetries, {\em non-equicardinal symmetries} in multi-valued domains, in which two variables that have different domain sizes may be symmetric. Our key insight is the framework of symmetry groups over variable-value (VV) pairs, instead of just variables. It allows interchanging a specific value of a variable with a different value of a different variable. 

Orbital MCMC suffices for downstream inference over most kinds of symmetries except non-equicardinal ones, for which a Metropolis Hastings extension is needed. Our new symmetries lead to substantial computational gains over Orbital MCMC and vanilla Gibbs Sampling, which doesn't exploit any symmetries. We make the following contributions:
\begin{enumerate}
\item We develop a novel framework for symmetries between variable-value (VV) pairs, which generalize existing notions of variable symmetries (Section \ref{sec:vv}).




\item We develop an extension of this framework, which can also identify Non-Equicardinal (NEC) symmetries, i.e., among variables of different cardinalities (Section \ref{sec:nonequi}).

\item We design a Metropolis Hastings version of Orbital MCMC called \necorbital\ to exploit NEC symmetries (Section \ref{sec:mcmc}). 

\item We experimentally show that our proposed algorithms significantly outperform strong baseline algorithms (Section \ref{sec:expt}). We also release the code for wider use\footnote{https://github.com/dair-iitd/nc-mcmc}.

\end{enumerate}

\section{Background}
\label{sec:background}
\vspace{-0.1in}
Let $\mathcal{X}=\{X_1,X_2,\cdots,X_n\}$ denote a set of Boolean valued variables. A state $s=\{(X_i,v_i)\}_{i=1}^n$ is a complete assignment to variables in $\mathcal{X}$, with values $v_i\in\{0,1\}$. We will use the symbol $\mathcal{S}$ to denote the entire state space.

A permutation $\theta$ of $\mathcal{X}$ is a bijection of the set $\mathcal{X}$ onto itself. $\theta(X_i)$ denotes the application of $\theta$ on the variable $X_i$. We will refer to $\theta$ as a {\em variable permutation}.  A permutation $\theta$ applies on state $s$ to produce $\theta(s)$, the state obtained by permuting the value of each variable $X_i$ in $s$ to that of $\theta(X_i)$. A set of permutations $\Theta$ is called a permutation group if it is closed under composition, contains the identity permutation, and each $\theta \in \Theta$ has its inverse in the set. 

A graphical model $\mathcal{G}$ over the set of variables $\mathcal{X}$ is defined as the set of pairs $\{f_j,w_j\}_{j=1}^{m}$ where $f_j$ is a feature function over a subset of variables in $\mathcal{X}$ and $w_j$ is the corresponding weight~\cite{koller&friedman09}. Drawing parallels from automporphism of a graph where a variable permutation maps the graph back to itself, we define the notion of automorphism (referred to as symmetry, henceforth) of a graphical model as follows~\cite{niepert13}.
\begin{defn}
A permutation $\theta$ of $\mathcal{X}$ is a {\em variable symmetry} of $\mathcal{G}$ if application of $\theta$ on $\mathcal{X}$ results back in $G$ itself, i.e., the same set $\{f_j,w_j\}_{j=1}^{m}$ as in $\mathcal{G}$. We also call such permutations as variable permutations.
\end{defn}

Correspondingly, we define the autormorphism group of a graphical model.
\begin{defn}
An {\em automorphism group} of a graphical model $\mathcal{G}$ is a permutation group $\Theta$ such that $\forall \theta \in \Theta$, $\theta$ is a variable symmetry of $\mathcal{G}$.
\end{defn}
Another important concept is the notion of an orbit of a state resulting from the application of a permutation group. 
\begin{defn}
The orbit ($\Gamma$) of a state $s$ under the permutation group $\Theta$, denoted by $\Gamma_{\Theta}(s)$, is the set of states resulting from application of permutations
$\theta \in \Theta$ on $s$, i.e., $\Gamma_{\Theta}(s)=\{s'| \exists \theta \in \Theta, \theta(s)=s'\}$.
\end{defn}

Note that orbits 
form an equivalence partition of the entire state space. In this work, we are interested in orbits obtained by application of an automorphism group, because all states in such an orbit have the same joint probability. Let $P_\G(s)$ denote the joint probability of a state $s$ under $\G$. 

\begin{thm}
Let $\Theta$ be an automorphism group of $\G$. Then for all states $s$ and permutations $\theta\in\Theta$: $P_\G(s) = P_\G(\theta(s))$.
\end{thm}

\subsection{Graph Isomorphism for Computing Symmetries}
\label{sec:isomorph}

The procedure for computing an automorphism group \cite{niepert12} first constructs a colored graph $G_V(\G)$ from the graphical model $\G$, in which all features are clausal or all features are conjunctive.\footnote{Each model can be pre-converted to a new model in which all features are clausal.} In this graph there are two nodes for each variable, one for each literal, and a node for each feature in $\G$. There is an edge between two literal nodes of a variable, and between a literal node and a feature if that literal appears in that feature in the graphical model. 
Each node is assigned a color such that all 1 value nodes get the same color, all 0 value nodes get the same color (but different from 1 node color), and all feature nodes get a unique color based on their weight. That is, two feature nodes have the same color if their weights in $\G$ are the same.

A graph isomorphism solver (e.g., Saucy \cite{saucy}) over $G_V(\G)$ outputs the automorphism group of this graph through a set of permutations. These permutations can be easily converted to variable permutations of $\G$, because any output permutation always maps a variable's 0 and 1 nodes to another variable's 0 and 1 nodes, respectively. These permutations collectively represent an automorphism group of $\G$.


\subsection{Orbital Markov Chain Monte Carlo}
\label{sec:orbitalmcmc}
Markov Chain Monte Carlo (MCMC) methods are one of most popular methods for inference where exact inference is hard. In these methods, a Markov chain $\mathcal{M}$ is set up over the state space and samples are generated.  Running the chain for a sufficiently long time, starts generating samples from the true distribution.
Gibbs sampling is one of the simplest MCMC methods. 

Orbital MCMC \cite{niepert12} adapts MCMC to use the given variable symmetries of the graphical model $\G$. 
Given a Markov Chain $\mathcal{M}$ and starting from state $s_t$, Orbital MCMC generates the next sample $s_{t+1}$ in two steps:
\begin{itemize}
\vspace{-1ex}
\item It first generates an intermediate state $s'_{t}$ by sampling from the transition distribution of $\mathcal{M}$ starting from $s_t$
\vspace{-1ex}
\item It then samples state $s_{t+1}$ uniformly from $\Gamma_{\Theta}(s'_t)$, the orbit of $s'_t$
\end{itemize}
The Orbital MCMC chain so constructed converges to the same stationary distribution as original chain $\mathcal{M}$ and is proven to mix faster, because of the orbital moves.
\section{Variable-Value (VV) Symmetries}
\label{sec:vv}
\vspace{-0.1in}
Existing work has defined symmetries in terms of variable permutations. We observe that these can only represent orbits in which all states have exactly the same count of 0s and 1s. The simple reason is that any variable permutation only permutes the values in a state and hence the total count of each value remains the same.  We name such type of symmetries as {\em count symmetries}. 

We now give a formal definition of count symmetries for a general multi-valued graphical model, since our work applies equally to both Boolean-valued as well as any other discrete valued domains. 
Let $\mathcal{X}=\{X_1,X_2,\cdots,X_n\}$ denote a set of variables where each $X_i$ takes values from a discrete valued domain $D_i$. A permutation $\theta$ of $\mathcal{X}$ is a {\em valid variable permutation} if it defines a mapping between variables having the same domain. 
Analogously, we define a {\em valid variable symmetry}. We will say that two domains $D_i$ and $D_j$ are {\em equicardinal} if $|D_i|=|D_j|$. We call such variables equicardinal variables. 
\begin{defn}
Given a set of variables $X \subseteq \mathcal{X}$ sharing the same domain $D$ and a $v \in D$, $count_{X}(s,v)$ computes the number of variables in $X$ taking the value $v$ in state $s$.
\end{defn}

\begin{defn}
Given a domain $D$, let $X_D$ denote the subset of all the variables whose domain is $D$. A (valid) variable symmetry $\theta$ is a {\em count symmetry} if for each such subset $X_D \subseteq \mathcal{X}$, $count_{X_D}(s,v)=count_{X_D}(\theta(s),v)$, $\forall v \in D, \forall s\in\mathcal{S}$.
\end{defn}


\begin{thm}
For a graphical model $\G$, every (valid) variable symmetry $\theta$ is a count symmetry.
\end{thm}

We argue here that count symmetries are restrictive; a lot more symmetry can be exploited if we simultaneously look at the {\em values} taken by the variables in a state. To illustrate this, consider a very simple graphical model  $\G_1$ with the following two formulas: (a) $w_1$: $a$ $\vee$ $\neg b$ (b) $w_2$: $\neg a$ $\vee$ $b$. 
It is easy to see that there is no non-trivial symmetry here. The permutation $\theta(a)=b,\theta(b)=a$ results in a different graphical model since the two formulas have different weights. On the other hand, if we somehow could permute $a$ with $\neg b$ and $b$ with $\neg a$, we would get back the same model. 
In this section, we will formalize this extended notion of symmetry which we refer to as {\em variable-value symmetry} (VV symmetry in short). 


\begin{defn}
Given a set of variables $\mathcal{X}=\{X_1,\cdots,X_n\}$ where each $X_i$ takes values from a domain $D_i$, a {\em variable-value} (VV) set is a set of pairs $\{(X_i,v^{i}_l)\}$ such that each variable $X_i$ appears exactly once with each $v^{i}_l \in D_i$ in this set where $v^{i}_l$ denotes the $l^{th}$ value in $D_i$. We will use  $\mathcal{S_X}$ to denote the VV set corresponding to $\mathcal{X}$.
\end{defn}
For example, given a set $\mathcal{X}=\{a,b\}$ of Boolean variables, the VV set is given by $\{(a,0),(a,1),(b,0),(b,1)\}$. 
\begin{defn}
A {\em Variable-Value permutation} $\phi$ over the VV set $\mathcal{S_X}$ is a bijection from $\mathcal{S_X}$  onto itself. 
\end{defn}

Recall that a variable permutation applied to a state in a Boolean domain always results in a valid state. However, that may not be true in multi-valued domains, since if two variables that have different domains are permuted, it may not result in a valid state. It is also not true for all VV permutations. For example, given the state $[(a,0),(b,0)]$, a VV permutation defined as $\phi(a,0)=(b,1),\phi(a,1)=(a,1)$, $\phi(b,0)=(b,0),\phi(b,1)=(a,0)$ results in the state [(b,1), (b,0)] which is inconsistent. Therefore, we need to impose a restriction on the set of allowed VV permutations so that they result in only valid states. 



\begin{defn}
We say that a VV permutation $\phi$ is a valid VV permutation if each variable $X_i \in \mathcal{X}$ maps to a unique variable $X_j$ under $\phi$. In other words, $\phi$ is valid if, whenever $\phi(X_i,v^{i}_{l})=(X_j,v^{j}_{l'})$ and  $\phi(X_i,v^i_{t})=(X_k,v^{k}_{t'})$, then $X_j=X_k$, $\forall v^i_{l},v^i_{t} \in D_i$. In such a scenario, we say that $\phi$ maps variable $X_i$ to $X_j$.
\end{defn}


It is easy to see that for any valid VV permutation $\phi$, applying $\phi$ on a state $s$ always results in a valid state $\phi(s)$. It also follows that if such a $\phi$ maps a variable $X_i$ to $X_j$, then $D_i$ and $D_j$ must be equicardinal.


\begin{thm}
The set of all valid VV permutations over $\mathcal{S_X}$ forms a group.
\end{thm}

Consider a graphical model $\mathcal{G}$ specified as a set of pairs $\{f_j,w_j\}$. Each feature $f_j$ can be thought of as a Boolean function over the variable assignments of the form $X_i=v^{i}_l$. Hence, action of a VV permutation $\phi$ on a feature $f_j$ results in a new feature $f_j'$ (with weight $w_j$) obtained by replacing the assignment $X_i=v^{i}_l$ by $X_j=v^j_{l'}$ in the underlying functional form of $f_j$ where $\phi(X_i,v^i_{l})=(X_j,v^j_{l'})$. Hence, application of $\phi$ on a graphical model $\mathcal{G}$ results in a new graphical model $\mathcal{G'}$ where each feature $(w_j,f_j)$ is transformed through application of $\phi$. We are now ready to define the symmetry of a graphical model under the application of VV permutations.
\begin{defn}
We say that a (valid) VV permutation is a {\em VV symmetry} of a graphical model $\mathcal{G}$ if application of $\phi$ on $\mathcal{G}$ results back in $\mathcal{G}$ itself.
\end{defn}

All other definitions of the previous section follow analogously. We can define an automorphism group over VV permutations, and also define an orbit of a state under this permutation group. 
VV symmetries strictly generalize the notion of variable symmetries. 

\begin{thm}

Each (valid) variable symmetry $\theta$ can be represented as a VV symmetry $\phi$. There exist valid VV symmetries that cannot be represented as a variable symmetry.
\end{thm}

Recall that a variable permutation $\theta$ is {\em valid} if it always maps between variables that have exactly the same domain. Say, $\theta(X_i) = X'_i$ with both variables having domains $D_i$. It is easy to see that $\phi$ defined such that $\phi(X_i, v^i_l)=(X'_i,v^i_l)$ for all $v^i_l\in D_i$, will result in the same sets of symmetric states.

To prove the second part, consider a PGM $\G_2$ with two Boolean variables $X_1$ and $X_2$. Let there be four features $f_{00},f_{01},f_{10},f_{11}$, one corresponding to each of the four states, with weights given as $w_s,w_d,w_d,w_s$, respectively.
Then, we have a VV symmetry $\phi$ such that $\phi(X_1,0)=(X_2,1)$, $\phi(X_1,1)=(X_2,0)$, $\phi(X_2,0)=(X_1,1)$ and $\phi(X_2,1)=(X_1,0)$. Note that $\phi$ maps the state $[(X_1,0),(X_2,0)]$ to $[(X_1,1),(X_2,1)]$ and reverse, and similarly there is a symmetry $\phi'$ which maps $[(X_1,0),(X_2,1)]$ to $[(X_1,1),(X_2,0)]$ and reverse. %
There is no variable symmetry which can capture the symmetries induced by $\phi$ since counts are not preserved. This proves the theorem. But let us for a moment define a renaming of the form $X'_1=\neg X_1$. Variable symmetries will now be able to capture the symmetries due to $\phi$ but will miss out on the ones due to $\phi'$. This is illustrative because there is no single problem formulation which can capture both the state symmetries above using the notion of variable symmetries alone.

\begin{thm}
VV symmetries preserve joint probabilities, i.e., for any VV symmetry $\phi$, and state $s$:  $P_\G(s)=P_\G(\phi(s))$.
\end{thm}

\begin{figure}

\includegraphics[width=0.23\textwidth]{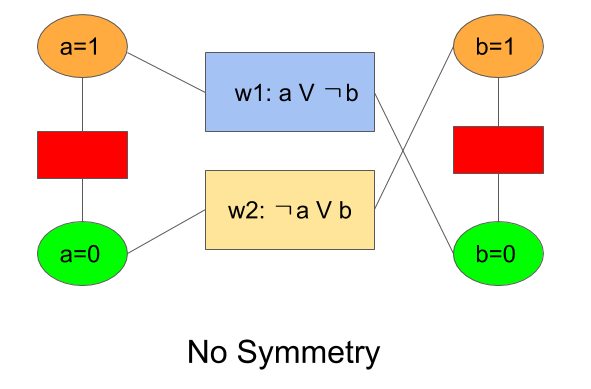}
{\includegraphics[width=0.23\textwidth]{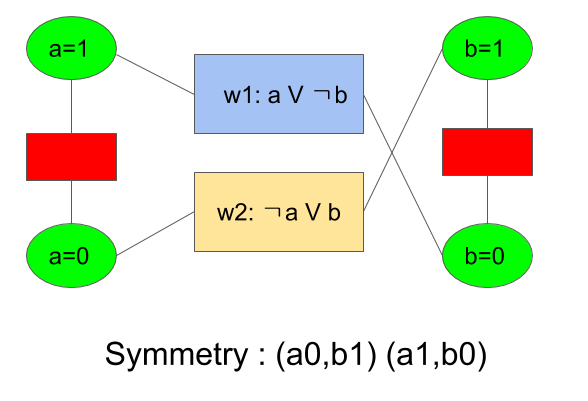}\label{fig:skill_one_sided}}
\vspace{-2ex}
\caption{{ {(a)} Variable Symmetry Graph for toy example $\G_1$} {{(b)} \vv \, Graph for $\G_1$}}
\label{fig:ex-graph}
\vspace{-2ex}
\end{figure}

\vspace{-0.1in}
\subsection{Computing Variable-Value Symmetries}
\label{sec:isomorph2}
We now adapt the procedure in Section \ref{sec:isomorph} to compute VV symmetries in multi-valued domains. For a PGM $\G$ with clausal theory or conjunctive theory (as before), we construct a colored graph $G_{VV}(\G)$ with a node for each variable-value pair. We also have a node for each feature, which is connected to the specific VV nodes it contains. We need to additionally impose a mutual exclusivity constraint to assert that a variable can only take exactly one of its many values. This is accomplished by adding exactly-one features with $\infty$ weight between all values of each variable. 
When assigning colors to each node, we assign all values of any variable the same color, as opposed to different values getting different colors. This allows the isomorphism solver to attempt discovering symmetries between different value nodes. As before, all features with the same weight get the same color. Figure \ref{fig:ex-graph} illustrates this on $\mathcal{G}_1$ where Variable symmetry assigns different colors to 0 and 1 while VV-Symmetry assigns a single color (green) to both 0 and 1 assignments of all variables.

We run Saucy \cite{saucy} over $G_{VV}(\G)$ to compute its automorphism group via a set of permutations. These permutations are valid VV-permutations (by construction of $G_VV$), and, 
collectively, represent a VV automorphism group of $\G$.


\begin{thm}
Any permutation $\phi$ that preserves graph isomorphism in $G_{VV}(\G)$ is a valid VV-permutation for $\G$.
\end{thm}

\begin{thm}
The automorphism group of colored graph $G_{VV}(\G)$ constructed above computes a VV-automorphism group of graphical model $\G$.
\end{thm}



\section{Non-Equicardinal (NEC) Symmetries}
\label{sec:nonequi}
\vspace{-0.1in}
\begin{figure}
\begin{center}
{\includegraphics[width=0.35\textwidth]{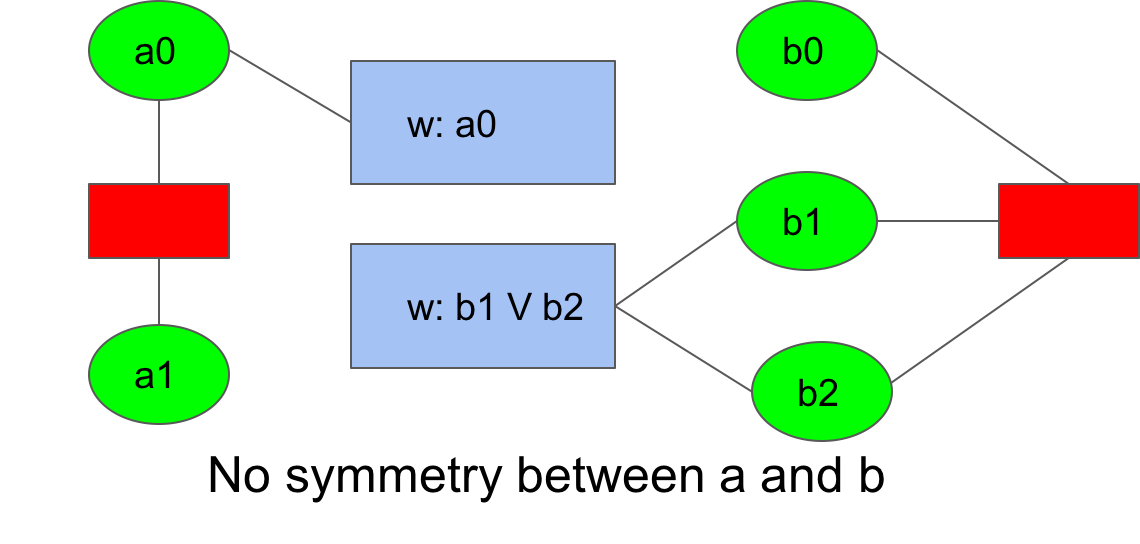}\label{fig:unreduced}}
{\includegraphics[width=0.35\textwidth]{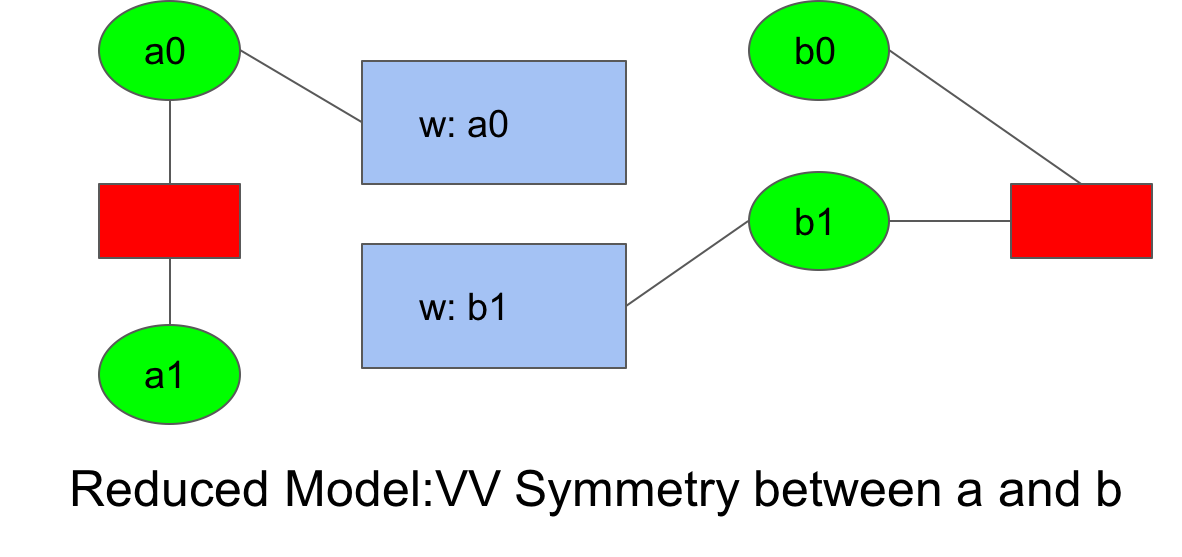}\label{fig:reduced}}
\end{center}
\vspace{-2ex}
\caption{{ (a)Unreduced multi-valued domain $\G_3$}  { (b) Reduced multi-valued domain $\G_3$}}
\label{fig:ex-reduction}
\vspace{-2ex}
\end{figure}

While VV symmetries can compute non-count symmetries, they only consider mapping between equicardinal variables.
In this section, we will deal with symmetries which can be present across variables having different domain sizes. Consider the following example graphical model $\G_3$ with two features: (1)$w$: $a=1$ (2) $w$: $b=1 \vee b=2$. Let $a$ and $b$ have the domains $D_a$ and $D_b$, respectively, specified as $D_a=\{0,1\}$ and $D_b=\{0,1,2\}$. Clearly, there is no VV symmetry between $a$ and $b$ since they have different domain sizes. But intuitively, the two states given as $[(a,1),(b,0)]$ and $[(a,0),(b,1)]$ are symmetric to each other since in each case, exactly one of the two features having the same weight is satisfied. Similarly, for $[(a,1),(b,0)]$ and $[(a,0),(b,2)]$. Further, it is easy to see that the two values of $b=1$ and $b=2$ are symmetric to each other in the sense states of the form $[(a,v),(b,1)]$ have the same probability as the states $[(a,v),(b,2)]$ where $v \in \{0,1\}$.

We will combine the above two ideas together to exploit symmetries using domain reduction. We first identify all the equivalent values of each variable and {\it replace} them by a single representative value. In this reduced graphical model, we then identify VV symmetries and finally translate them back to the original graphical model. In the following, we will assume that we are given a graphical model $\mathcal{G}$ defined over a set of $n$ variables $\mathcal{X}$ where each $X_i \in \mathcal{X}$ takes values from a domain $D_i$. Further, we will use the symbol $\mathcal{D}=D_1 \times D_2,\cdots D_n$ to denote the cross product of the domains. 

\begin{defn}
Consider a variable $X_i \in\mathcal{X}$ and let $v,v' \in D_i$. Let $\phi^i_{v \leftrightarrow v'}$ denote a VV permutation which maps the VV pair $(X_i,v)$ to $(X_i,v')$ and back.
For all the remaining VV pairs $(X_k,v'')$, $\phi^i_{v \leftrightarrow v'}$ maps the pair back to itself. We refer to $\phi^i_{v \leftrightarrow v'}$ as a {\em value swap permutation} for variable $X_i$.
\end{defn}
In the example above, $\phi^b_{1 \leftrightarrow 2}$ is a value swap permutation for $b$ which permutes the variable assignments $b=1$ and $b=2$, and keeps the remaining variable assignments, i.e., $b=0$ and $a=1$, fixed.
\begin{defn}
A value swap permutation $\phi^i_{v \leftrightarrow v'}$ is a {\em a value swap symmetry} of $\mathcal{G}$ if it maps $\mathcal{G}$ back to itself.
\end{defn}
In our running example, $\phi^b_{1 \leftrightarrow 2}$ is a value swap symmetry of $\G_4$. Next, we show that the set of all value swap symmetries corresponding to a variable $X_i$ divides its domain $D_i$ into equivalence classes.
\begin{defn}
Given a graphical model $\mathcal{G}$, we define a relation $SS_i$ (swap symmetry) over the set $D_i \times D_i$ as follows. Given $v,v' \in D_i$, $(v,v') \in SS_i$ if $\phi^i_{v \leftrightarrow v'}$ is a value swap symmetry of $\mathcal{G}$. 
\end{defn}
It is easy to see that relation $SS_i$ is an equivalence relation and hence, partitions the domain $D_i$ into a set of equivalence classes. Given a value $v \in D_i$, we choose a representative value from its equivalence class based on some canonical ordering. We denote this value by $rep_i(v)$.

Next, we will define a reduced domain $D_{i}^{R}$ obtained by considering one value from each equivalence set. 
\begin{defn}
Let $SS_i$ divide the domain $D_i$ into $r$ equivalence classes. We define the reduced domain $D_i^{R}$ as the $r$-sized set $\{v^*_{j}\}_{j=1}^{r}$ where $v^*_{j}$ is the representative value for the $j^{th}$ equivalence class. We will use $\mathcal{D}^R=D^R_1\times D^R_2\times \cdots D^R_n$ to denote the cross product of the reduced domains.
\end{defn}
Revisiting our example, the reduced domain for $b$ is given as $D_b^{R}=\{0,1\}$. Next we define a reduced graphical model $\mathcal{G}^R$ over the reduced set of domains $\{D_i^{R}\}_{i=1}^{n}$. 
\begin{defn}
Let $\mathcal{G}$ be a graphical model with the set of weighted features $\{w_j,f_j\}$. 
Let $X_i=v$ be a variable assignment appearing in the Boolean expression for $f_j$. We construct a new feature $f'_j$ by replacing every such expression $X_i=v$ by $false$ (and further simplifying the expression) whenever $v \neq rep_i(v)$. If $v=rep_i(v)$, then we leave the assignment $X_i=v$ in $f'_j$ as is. The reduced $\mathcal{G}^R$ is the graphical model having the set of features $\{w_j,f'_j\}$ defined over the set of variables $\mathcal{X}$ with $X_i$ having the domain $D^R_i$. 
\end{defn}
Intuitively, in $\G^{R}$, we restrict each variable $X_i$ to take only the representative value from each of its equivalence classes. In our running example, the reduced graphical model is given as \{$w$: $a=0$; $w$: $(b=1) \vee false$\} which is same as \{$w$: $a=0$; $w$: $b=1$\}. Since the domains have been reduced in $\mathcal{G}^R$, we may now be able to discover mappings which were not possible earlier. For instance in our running example, we now have a VV symmetry $\phi^R$ which maps $(a,0)$ to $(b,1)$ and back. 

Let the joint distributions specified by $\mathcal{G}$ and $\mathcal{G}^R$ be given by $P_{G}$ and $P_{G^R}$, respectively. The next theorem describes the relationship between these two distributions.
\begin{thm}
\label{thm:prob_equ}
Let $\mathcal{G}$ be a graphical model and let $\mathcal{G}^R$ be the corresponding reduced graphical model.  Consider a state $s$ specified as $\{X_i,v_i\}_{i=1}^{n}$
where each $v_i \in D^{R}_i$. By definition, $v_i \in D_i$. We claim that $P_{\mathcal{G}^R}(s)=k*P_{\mathcal{G}}(s)$ where $k$ is some constant $k \ge 1$ independent of the specific state $s$.
\end{thm}
\vspace{-0.2in}
\begin{proof}
Note that the reduced graphical model $G^R$ is emulating the distribution specified by $\mathcal{G}$ where the space of possible variable assignments is now restricted to those 
belonging to the representative set, i.e., for each variable $X_i$ the allowed set of values is now $D^R_i=\{v_i|v_i=rep_i(v), v\in D_i\}$. Therefore, $G^R$ can be thought of as enforcing a conditional distribution over the underlying space given the fact that assignments can now only come from cross product set $\mathcal{D}^R$. Recall that state $s$
is valid assignment in the original as well as the reduced graphical model. 
Therefore, we have $P_{\mathcal{G}^R}(s)=P_{\mathcal{G}}(s|s \in \mathcal{D}^R)=P_{\mathcal{G}}(s)/P_{\mathcal{G}}(s \in \mathcal{D}^R)$. Here, the denominator term $P_{\mathcal{G}}(s \in \mathcal{D}^R)$ is simply the probability that a randomly chosen state $s$ in the original distribution belongs to the restricted domain set. Clearly, this is independent of the state $s$ and let this given as $1/k$, where $k \ge 1$ is a constant independent of $s$. Then, $P_{\mathcal{G}^R}(s)=k*P_{\mathcal{G}}(s)$. 
\end{proof}
Above theorem gives us a recipe to discover additional symmetries across variables having different domain sizes. Let $s=\{X_i,v_i\}_{i=1}^{n}$ 
be a state in $\mathcal{G}$. Let $rep(s)$ denote the representative state for $s$ given as
$\{(X_i,rep_i(v_i)\}_{i=1}^{n})$.
Following steps describe a procedure to get a new state $s'$ symmetric to $s$ using the idea of domain reduction.\\\\
{\bf Procedure NonEquiCardinalSym:}
\begin{itemize}
\vspace{-0.1in}
\item Let $u=rep(s)$ denote the representative state for $s$.
\vspace{-1ex}
\item Apply a VV symmetry $\phi^{R}(u)$ over $u$ in the reduced graphical model. Resulting state $u'$ is symmetric to u in $\mathcal{G}^R$. 
\vspace{-1ex}
\item Apply a series of $n$ value swap symmetries of the form $\phi^{i}_{v'_i \leftrightarrow v''_i}$ over state $u'$, one for each variable $X_i$ such that $X_i=v'_i$ in $u'$, $v''_i \in D_i$. Resulting state $s'$ is  symmetric to $s$ in $\mathcal{G}$.
\end{itemize}
\begin{defn}
Let $\tau$ be a permutation over the state space $\mathcal{S}$ of $\mathcal{G}$
defined using the Procedure NonEquiCardinalSym, i.e., $\tau(s)=\phi^n_{v'_n \leftrightarrow v''_{n}}(\phi^{n-1}_{v_{n-1} \leftrightarrow v''_{n-1}}( \cdots \phi^1_{v'_1 \leftrightarrow v''_1}(\phi^{R}(rep(s))) \cdots ))$, where  $\phi^R$ is a VV symmetry of $\mathcal{G}^R$ and each $\phi^i_{v'_i\leftrightarrow v''_i}$ is a value swap symmetry for variable $X_i$ in $\mathcal{G}$ . We refer to $\tau$ as a non-equicardinal symmetry of $\mathcal{G}$.
\end{defn}
Unlike VV symmetries whose action is defined over a VV pair, non-equicardinal symmetries directly operate over the state space. Their transformation of the underlying graphical model is implicit in the symmetries that compose them.
\begin{thm}
The set of all non-equicardinal symmetries forms a permutation group. 
\end{thm}
Finally, we need to show that action of non-equicardinal symmetries indeed results in states which have the same probability.
\begin{thm}
Let $\tau$ be a non-equicardinal symmetry of a graphical model $\mathcal{G}$. Then, $P_\mathcal{G}(s)=P_\mathcal{G}(\tau(s))$.
\end{thm}
\vspace{-0.2in}
\begin{proof}
Let $s'=\tau(s)$. Let $u=rep(s)$. Since $u'$ is obtained by application of VV symmetry $\phi^R(u)$ in $\mathcal{G}^R$, we have $P_\mathcal{G}^R(u)=P_\mathcal{G}^R(u')$. Using Theorem~\ref{thm:prob_equ}, this implies that $P_{\mathcal{G}}(u)=(1/k)*P_{\mathcal{G}^R}(u)=(1/k)*P_{\mathcal{G}^R}(u')=P_{\mathcal{G}}(u')$ for some constant $k$. Hence, $u$ and $u'$ have the same probability under $P_{\mathcal{G}}$. 

Since $u=rep(s)$ can be obtained by application of $n$ value swap symmetries over $s$ (one for each variable), $P_{\mathcal{G}}(s)=P_{\mathcal{G}}(u)$. Similarly, since $s'$ is obtained by an application of $n$ value swap symmetries over $u'$, we have $P_{\mathcal{G}}(u')=P_{\mathcal{G}}(s)$. Combining this with the fact that, $P_{\mathcal{G}}(u)=P_{\mathcal{G}}(u')$, we get $P_{\mathcal{G}}(s)=P_{\mathcal{G}}(s')$. 
\end{proof}


\subsection{Computing Non-Equicardinal Symmetries}
We adapt the procedure in Section \ref{sec:isomorph2} by running graph isomorphism over a series of two colored graphs. Our first colored graph is constructed as in Section \ref{sec:isomorph2}, except that all features are given different colors. This disallows any mapping between $(X_i,v_i)$ and $(X_j,v_j$) for $X_i \neq X_j$, and only allows mapping between different values of a single variable. For example, in the running example, this would determine that $(b,1)$ and $(b,2)$ are symmetric. We then retain only the representative value for each equivalent set of VV pairs, and removes nodes and edges for other values. 

We take this reduced colored graph and recolor all mutual exclusivity features with a single color. We run graph isomorphism again to obtain the VV symmetries of the reduced model. These permutations together with the single-variable permutations from the previous step gives the non-equicardinal symmetries of the original model.


\section{MCMC with VV \& NEC Symmetries}
\label{sec:mcmc}
\vspace{-0.1in}
\begin{figure}
\begin{center}
{\includegraphics[width=0.35\textwidth]{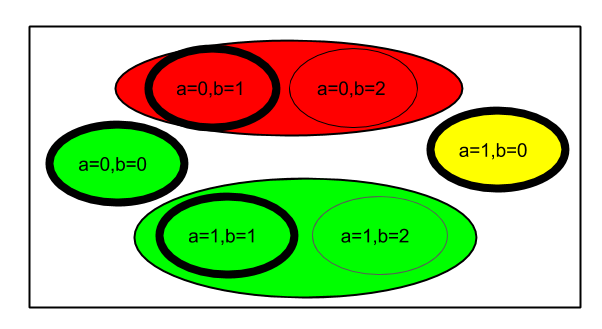}\label{fig:statePart}}
\end{center}
\vspace{-2ex}
\caption{State Partition for Toy Example $\G_3$. Same Colored States are in same orbit. Large Ovals show sub-orbits and representative states of sub-orbits are with dark outline. }
\vspace{-2ex}
\label{fig:states}

\end{figure}

\begin{figure*}
\begin{center}
\framebox{
\subfigure[]{
{\includegraphics[width=0.28\textwidth]{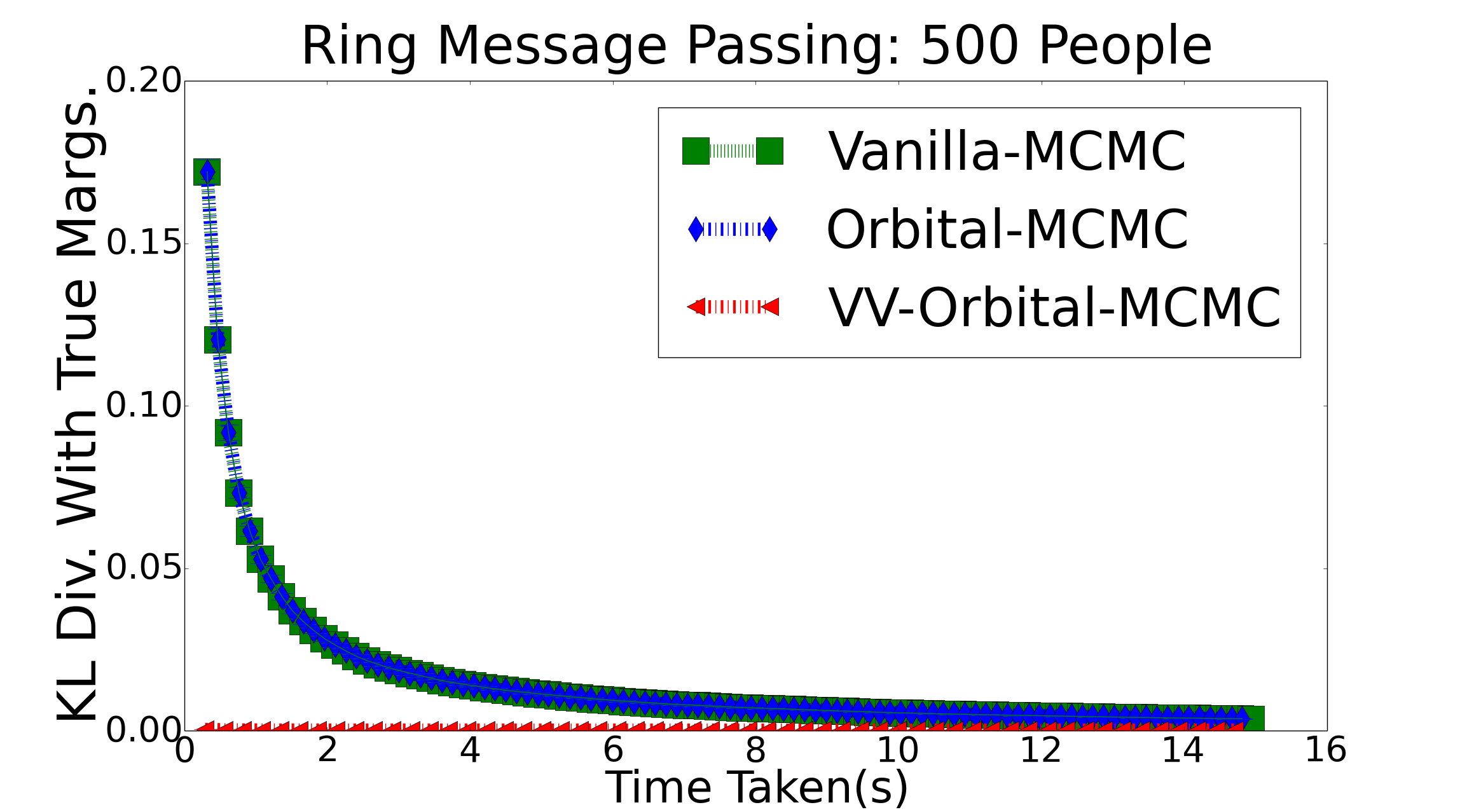}\label{fig:table1}}
{\includegraphics[width=0.28\textwidth]{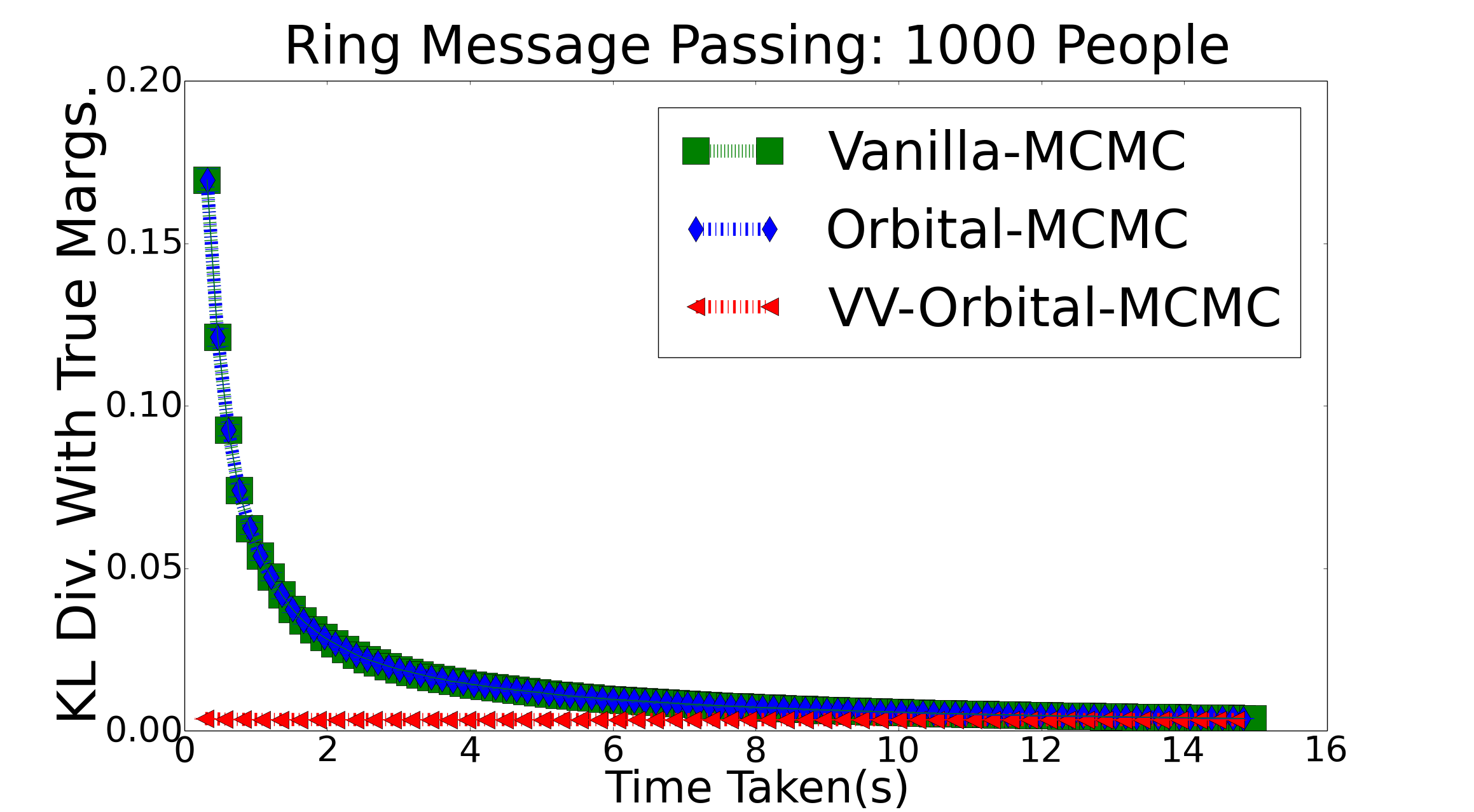}\label{fig:table2}}}\label{fig:table}}
\subfigure[]{
{\includegraphics[width=0.28\textwidth]{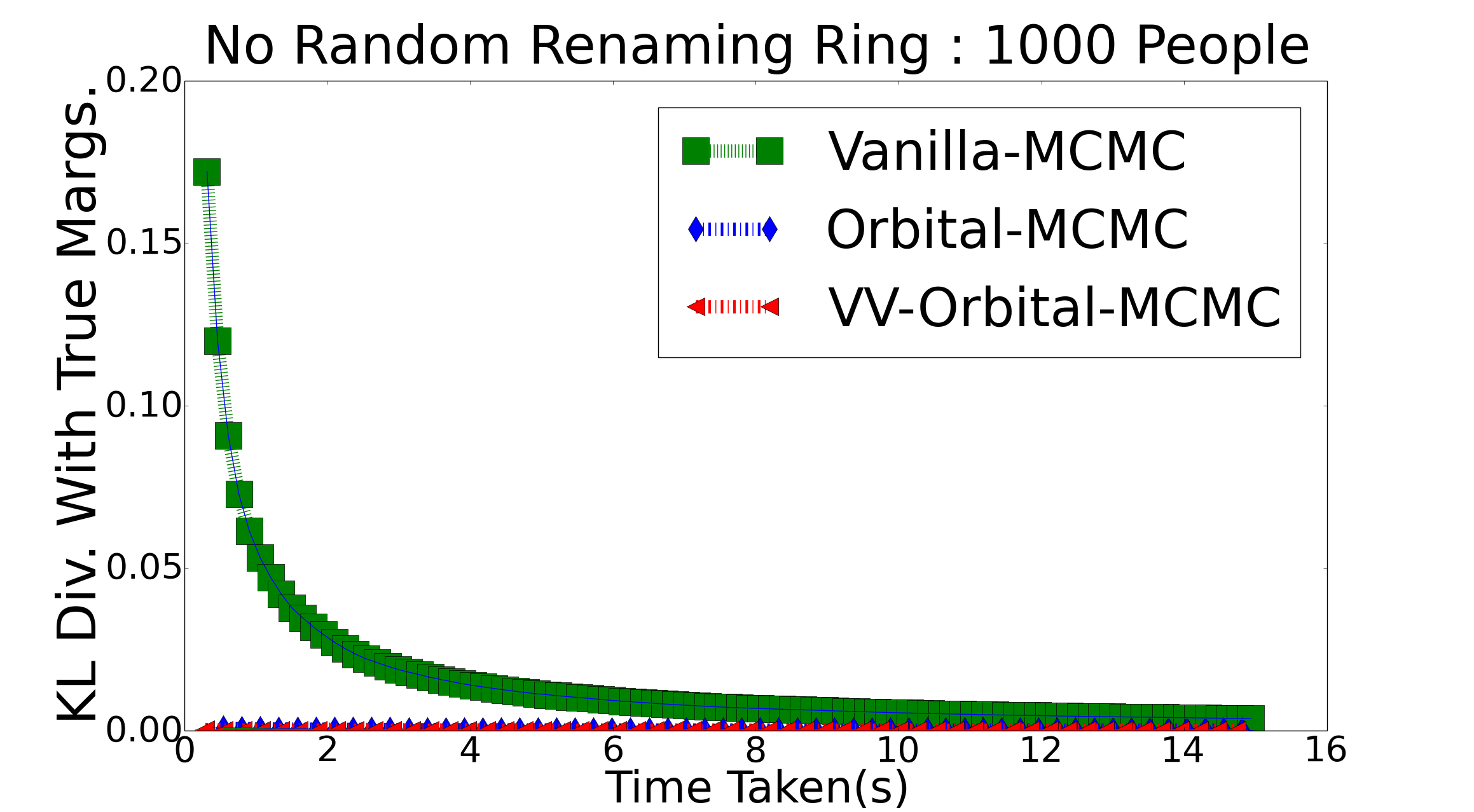}}\label{fig:ringEqual}}
\end{center}
\vspace{-2ex}
\caption{{a)}VV-Orbital-MCMC outperforms Orbital MCMC and Vanilla MCMC with different sizes of people on ring-message passing. {b)} \vvorbital\ has negligible overhead compared to Orbital-MCMC}
\vspace{-2ex}
\end{figure*}
Recall from Section \ref{sec:orbitalmcmc} that variable symmetries are used in approximate inference via the Orbital MCMC algorithm. It alternates original MCMC move with an orbital move, which uniformly samples from the orbit of the current state. We first observe that the same algorithm will work for VV symmetries computed in Section \ref{sec:isomorph2}, except that the orbital move will now sample from the orbit induced by VV permutations -- we call this algorithm \vvorbital.




We now consider the case of non-equicardinal symmetries in multi-valued PGMs. The main idea from Orbital MCMC remains valid -- we need to alternate between original chain and orbital move. However, sampling a random state from an orbit is tricky now, because a non-equicardinal orbit may have a two-level hierarchical structure -- it is an orbit over suborbits. The top level orbit is in the reduced model and is an orbit over representative states. At the bottom level, each representative state may represent multiple states via application of a {\em variable} number of value-swap symmetries. 

As an example, consider the state partition in our running example, as illustrated in Figure \ref{fig:states}. 
Each orbit is shown by a unique color, and suborbits by large ovals. The green orbit (top level) has two representative states (0,0) and (1,1) in the reduced model. If we make an orbital move in the reduced model, we can easily pick a representative state uniformly at random. However, the state (1,1) has a suborbit -- it further represents two states in the original model, (1,1) and (1,2), via value-swap symmetries on variable $b$. Our sampling goal is to pick uniformly at random from an orbit in the {\em original} model, which means we need to pick a representative state in the reduced model proportional to the size of suborbit it represents. Once a suborbit is picked, we can easily pick a state uniformly at random from within it. To pick a representative proportional to the size of the suborbit, we use Metropolis Hastings in the reduced model -- we name the resulting algorithm \necorbital. 

Let $c(s)$ represent the cardinality of the suborbit of state $s$, i.e., the number of states for which the representative state is the same as that of $s$: $|\{s'|rep(s')=rep(s)\}|$. Let $c^i(s)$ represent the number of states in the orbit of $s$ which differ from $s$ at most on the value of $X_i$ 
, i.e., $|\{s'|rep(s')=rep(s), s.X_j = s'.X_j \forall j\neq i\}|$, where $s.X_j$ represents the value of $X_j$ in $s$.

Given a Markov chain $\mathcal{M}$ over a graphical model $\G$,  a sample from $s_t$ to $s_{t+1}$ in \necorbital\ is generated: 
\begin{itemize}
\vspace{-1ex}
\item Generate $s'_t$ by sampling from transition distribution of $\mathcal{M}$ starting from $s_t$.
\vspace{-1ex}
\item Let $u'_t = rep(s'_t)$. Sample $u''_t$ (in $\mathcal{G}^R$) from the orbit $\Gamma_{\phi^R}(u'_t)$ via a Metropolis Hastings step using the uniform proposal distribution $q(\cdot)=\frac{1}{|\Gamma_{\Phi^{R}}(u'_t)|}$, and desired distribution $p(\cdot) \propto c(u''_t)$. 
\vspace{-1ex}
\item Apply a series of $n$ value swap symmetries of the form $\phi^i_{v''^i_t\leftrightarrow v^i_{t+1}}$ over state $u''_t$, one for each variable $X_i$, where $u''_t.X_i=v''^i_t$, and $v^i_{t+1}$ is chosen uniformly at random from the set of values equivalent with $v''^i_t$ with probability $\frac{1}{c^i(u''_t)}$. This is equivalent to sampling uniformly from the suborbit of $u''_t$. 
\end{itemize}

Notice that sampling from the proposal distribution (uniform) from an orbit is easily accomplished by 
Product Replacement Algorithm \cite{pak00}. MH accepts or rejects the sample with an Acceptance probability $A$, which can be computed by MH's detailed balance equation:
\begin{eqnarray*}
A(u'_t \to u''_t)
&=& min\left(1,\frac{p(u''_t)*q(u'_t | u''_t)}{p(u'_t)*q(u''_t | u'_t)}\right) \\
&=& min\left(1,\frac{p(u''_t)}{p(u'_t)}\right)
= min\left(1,\frac{c(u''_t)}{c(u'_t)}\right)
\end{eqnarray*}
The second equality above follows form the fact that $q(.)$ is a uniform proposal.

\begin{thm}
The Markov Chain constructed by \necorbital\ converges to the unique stationary distribution of original markov chain $\mathcal{M}$.
\end{thm}


\vspace{-0.05in}
\section{Experiments}
\label{sec:expt}
\vspace{-0.1in}
We empirically evaluate our extensions of Orbital MCMC for both Boolean and multi-valued PGMs. In both settings, we compare against the baselines of vanilla MCMC, and Orbital MCMC \cite{niepert12}. In all orbital algorithms including ours, the base Markov chain $\mathcal{M}$ is set to Gibbs. 
We build our source code on existing code of Orbital MCMC.\footnote{https://code.google.com/archive/p/lifted-mcmc/} It uses the Group Theory package Gap \cite{gap15} for implementing the group-theoretic operations in the algorithms. We release our implementations for further research. \footnote{Available at {\em https://github.com/dair-iitd/nc-mcmc}} All our experiments are performed on Intel core i-7 machine. All our reported times include the time taken for computing symmetries.

Our experiments are aimed to assess the comparative value of our algorithms against baselines in those domains where a large number of symmetries (beyond count symmetries) are present. To this end, we construct two such domains. The first is a simple Boolean domain that shows how simple value renaming can affect baseline algorithms. The second is a multi-valued domain showcasing the potential benefits of non-equicardinal symmetries. The domains are:

{\bf Value-Renamed Ring Message Passing Domain: } In this simple domain, $N$ people with equal number of males and females are placed in a ring structure alternately with every male followed by a female, and they pass a bit of message to their immediate neighbor over a noisy channel. If $X_i$ denoted the bit received by the $i^{th}$ person, then we would have a formula for PGM $X_i \rightarrow X_{i+1}$ with weight $w_1$ if $i$ is a male and weight $w_2$ if $i$ is female. As a small modification to this domain, we randomly rename some $X_i$s to mean $\neg$bit received by that agent, and change all formulas analogously. All the symmetries in the original ring should remain after this renaming. Our experiments test the degree to which the various algorithms are able to identify these.

{\bf Student-Curriculum Domain: } In this multi-valued domain, there are $K$ students taking courses from $|A|$ areas (e.g., theory, architecture, etc.). Each area $a\in A$ has a variable number of $N(a)$ courses numbered 1 to $N(a)$. Each student has to fulfill their breadth requirements by passing one course each from any two areas. A student has no specific preference to which of the $N(a)$ courses they take in an area. However, each student has a prior seriousness level, which determines whether they will pass {\em any} course.  This scenario is modeled by defining a random variable $P_{sa}$, which is a multi-valued variable where value $0$ denotes that student $s$ failed the course in the area $a$, and value $i\in \{1:N(a)\}$ denotes which course they passed. The weight for failing depends on the student but not on area. Finally, the variable $C_{saa'}$ denotes that $s$ completed their requirements by passing courses from areas $a$ and $a'$.

The Curriculum domain is interesting, because, for a given $s$,  various values of $P_{sa}$ other than 0 are all symmetric for all areas. And once, all $P_{sa}$s are converted to a representative value in the  reduced model, all areas become symmetric for a student.

\begin{figure}
\begin{center}
{\includegraphics[width=0.35\textwidth]{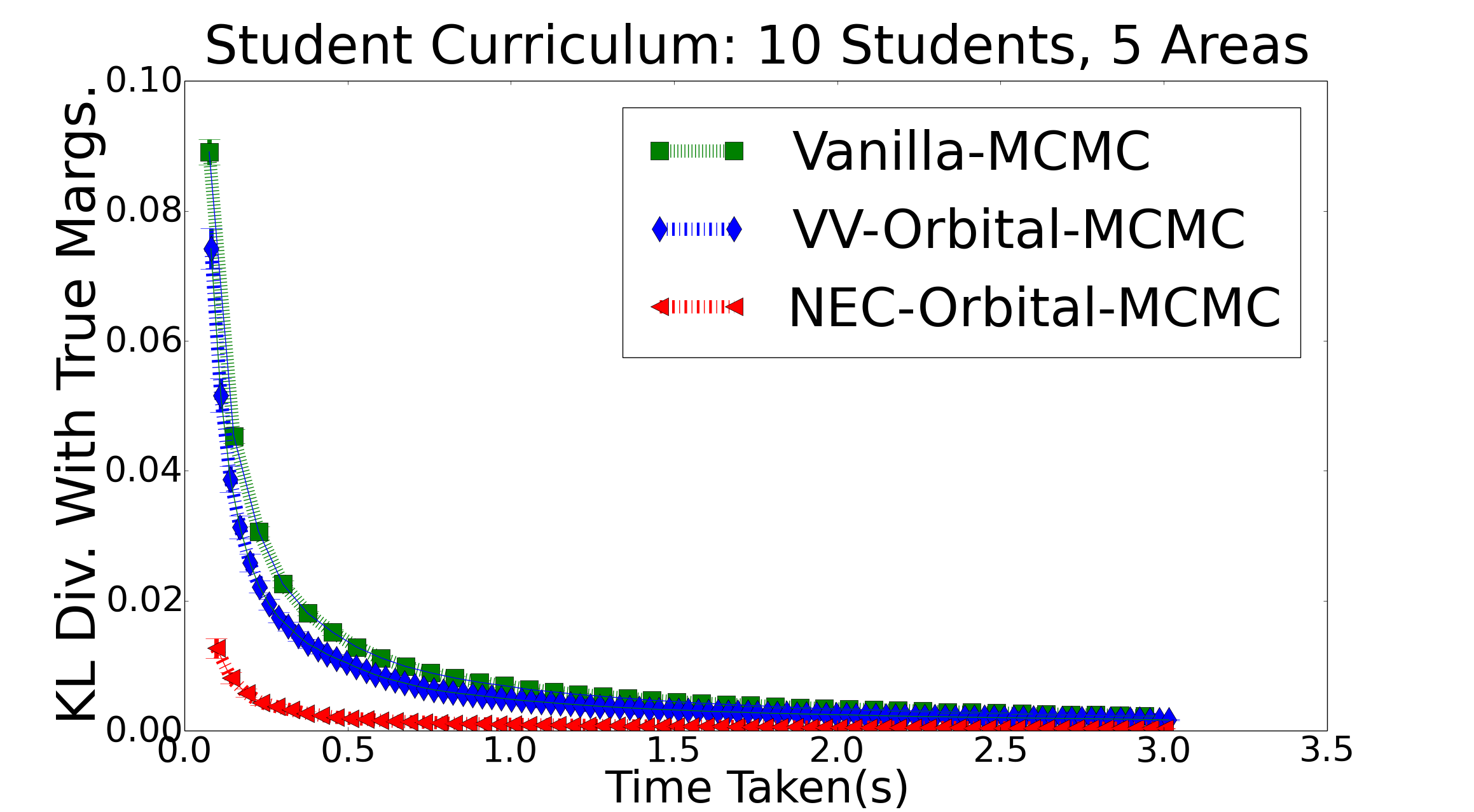}\label{fig:res1}}
{\includegraphics[width=0.35\textwidth]{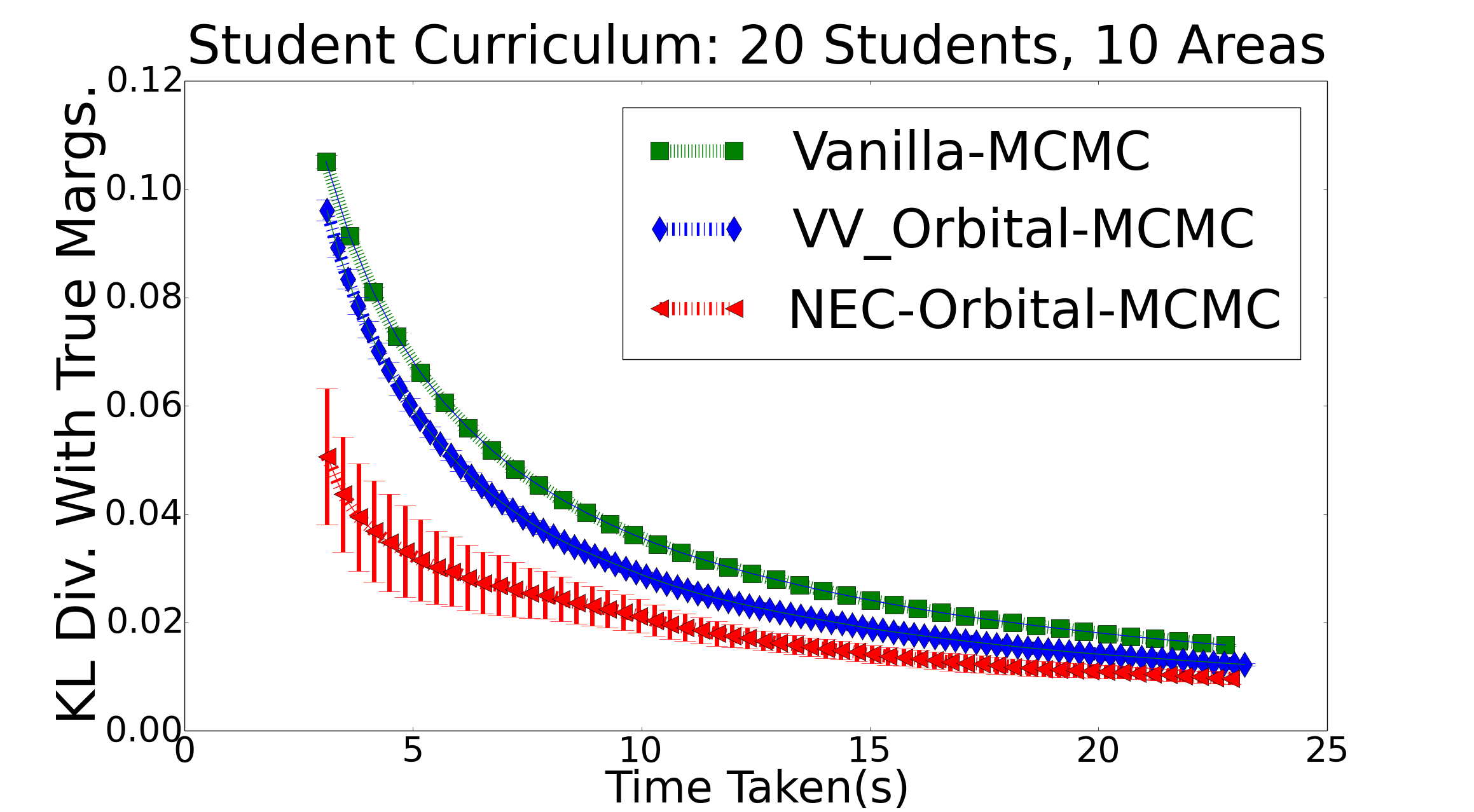}\label{fig:res2}}
\end{center}
\vspace{-2ex}
\caption{\necorbital\ outperforms \vvorbital\ and Vanilla-MCMC on student-curriculum domain.}
\label{fig:res}
\vspace{-2ex}
\end{figure}

We compare different algorithms by plotting the KL-divergence of true marginals and an algorithm's marginals with time. True marginals are calculated by running Gibbs sampling for a sufficiently large duration of time. Figure \ref{fig:table} compares \vvorbital\ with baselines on the message passing domain. The dramatic speedups obtained by \vvorbital\ underscores Orbital MCMC's inability to identify the huge number of variable-renamed symmetries present in this domain, whereas \vvorbital\ is able to benefit from these tremendously.

Before describing results on Curriculum domain, we first highlight that, out of the box, Orbital MCMC cannot run on this domain, because both its theory and implementation have only been developed for Boolean-valued PGMs. To meaningfully compare against Orbital MCMC, we first {\em binarize} the domain, by converting each multi-valued random variable $P_{sa}$ into many Boolean variables $P_{sac}$, one for each value $c$. We need to add an infinite-weighted exactly-one constraint for each original variable before giving it to Orbital MCMC. A careful reader may observe that this binarization is already very similar to the VV construction of Section \ref{sec:vv}, but without non-equicardinal symmetries. Thus, this is already a much stronger baseline than currently found in literature.

Figure \ref{fig:res} shows the results on this domain. \necorbital\ outperforms both baselines by wide margins. Orbital MCMC does improve upon vanilla Gibbs since it is able to find that all $P_{sac}$s for different $c$s are equivalent, however, it is unable to combine them across areas. 

In domains where symmetries beyond count symmetries are not found, the overhead of our algorithms is not significant, and they perform almost as well as (binarized) Orbital MCMC (e.g., see Figure \ref{fig:ringEqual}). This is also corroborated by the fact that the time for finding symmetries is relatively small compared to the time taken for actual inference on both the domains. Specifically, this time is 0.250 sec and 0.009 sec for curriculum and ring domains, respectively.

In summary, both \vvorbital\ and \necorbital\ are useful advances over Orbital MCMC.
\vspace{-0.1in}
\section{Conclusion and Future Directions}
\vspace{-0.1in}
Existing lifted inference algorithms capture only a restricted set of symmetries, which we define as {\em count symmetries}.  To the best of our knowledge, this is the first work that computes symmetries beyond count symmetries. To compute these {\em non-count symmetries}, we introduce the idea of computation over variable-value (VV) pairs. We develop a theory of VV automorphism groups, and provide an algorithm to compute these. These can compute equicardinal non-count symmetries, i.e., between variables that have the same cardinality. An extension to this allows us to also compute {\em non-equicardinal symmetries}. Finally, we provide MCMC procedures for using these computed symmetries for approximate inference. In particular, the algorithm to use non-equicardinal symmetries requires a novel Metropolis Hastings extension to existing Orbital MCMC. Experiments on two domains illustrate that exploiting these additional symmetries can provide a huge boost to convergence of MCMC algorithms. 

We believe that many real world settings exhibit VV symmetries. For example, in the standard Pott’s model used in Computer Vision \cite{koller&friedman09}, the energy function depends on whether the two neighboring particles take the same value or not, and not on the specific values themselves (hence, 00 would be symmetric to 11). Exploring VV symmetries in the context of specific applications is an important direction for future research. 

We will also work on extending the theoretical guarantees of variable symmetries \cite{niepert12} to VV symmetries. Several notions of symmetries already exist in the Constraint Satisfaction literature \cite{cohen&al06}. It will be interesting to see how our approach can be incorporated into the existing framework of symmetries in CSPs.

\newpage
\section*{Acknowledgements}
We thank Mathias Niepert for his help with the orbital-MCMC code. Ankit Anand is being supported by the TCS Research Scholars Program. Mausam is being supported by grants from Google and Bloomberg. Both Mausam and Parag Singla are being supported by the Visvesvaraya Young Faculty Fellowships by Govt. of India.
\bibliography{references}
\nocite{crawford&al96}
\nocite{kopp&al15}
\nocite{broeck&darwiche13}
\nocite{sarkhel&al14}
\nocite{mittal&al14}
\nocite{mladenov&al12}
\bibliographystyle{plain}
\end{document}